%% file: naacl2022.tex
\pdfoutput=1

\documentclass[11pt]{article}

\usepackage{acl}

\usepackage{times}
\usepackage{latexsym}

\usepackage[T1]{fontenc}

\usepackage[utf8]{inputenc}

\usepackage{microtype}

\usepackage{amsmath}
\usepackage{amssymb}
\usepackage{bbm}
\usepackage{graphicx} 
\usepackage{booktabs}
\usepackage{multirow}

\DeclareMathOperator*{\argmax}{arg\,max}

\title{Self-Aware Feedback-Based Self-Learning in Large-Scale\\Conversational AI}

\author{
    Pragaash Ponnusamy\thanks{\qquad Equal contribution}\qquad Clint Solomon Mathialagan\footnotemark[1]\\
    {\bf Gustavo Aguilar} \qquad {\bf Chengyuan Ma} \qquad {\bf Chenlei Guo} \\
    Amazon Alexa \\
    \texttt{\{ponnup,matclint,gustalas,mchengyu,guochenl\}@amazon.com} \\
}

\begin{document}
\maketitle

\begin{abstract}
Self-learning paradigms in large-scale conversational AI agents tend to leverage user feedback in bridging between what they say and what they mean. However, such learning, particularly in Markov-based query rewriting systems have far from addressed the impact of these models on future training where successive feedback is inevitably contingent on the rewrite itself, especially in a continually updating environment. 
In this paper, we explore the consequences of this inherent lack of self-awareness towards impairing the model performance, ultimately resulting in both Type I and II errors over time. 
To that end, we propose augmenting the Markov Graph construction with a superposition-based adjacency matrix. Here, our method leverages an induced stochasticity to reactively learn a locally-adaptive decision boundary based on the performance of the individual rewrites in a bi-variate beta setting. %
We also surface a data augmentation strategy that leverages template-based generation in abridging complex conversation hierarchies of dialogs so as to simplify the learning process. All in all, we demonstrate that our self-aware model improves the overall PR-AUC by 27.45\%, achieves a relative defect reduction of up to 31.22\%, and is able to adapt quicker to changes in global preferences across a large number of customers.
\end{abstract}

\input{sections/intro}
\begin{figure*}[h]
    \centering
    \includegraphics[width=\textwidth]{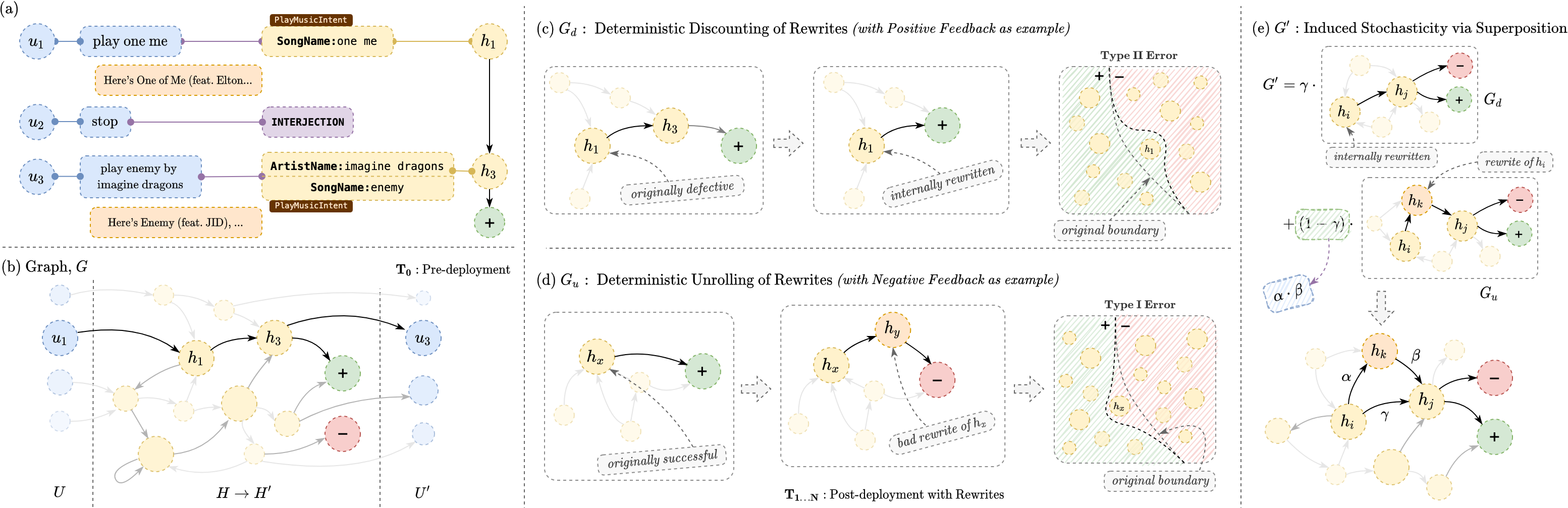}
    \caption{ 
        A general walk-through for motivating a meta-state augmented Graph: Beginning with the original construction of chains in \textbf{(a)} where utterances, $U$ are projected into the hypothesis space, $H$ before being encoded into the absorbing Markov model in \textbf{(b)} showing how a target rewrite in $U^{\prime}$ is resolved given a source in $U$. Thereafter, upon deployment, the effect of continuing to model the Graph as before i.e. by discounting the presence of rewrites, $G_d$ in \textbf{(c)} and choosing to always unroll the internal rewrites as an externalized state, $G_u$ in \textbf{(d)}, both lead to Type II and I errors respectively. Note that the decision boundaries over discrete spaces here are to illustrate the nature of mis-classifications. Naturally, in attempt to balance these two categories of error, a superposition of $G_d$ and $G_u$ is constructed in \textbf{(e)} wherein the rewrites act as meta-states that induce stochasticity within the Graph, $G^{\prime}$.
    }
    \label{fig:main}
\end{figure*}

\input{sections/related}

\input{sections/dataset}

\input{sections/model}

\input{sections/experiments}

\input{sections/deployment}

\section{Conclusion}
In this work, we address one of the key hurdles to the achieving self-learning in continuously updated feedback based systems, namely the deformation of the partitioning decision boundary due to lack of self-awareness.
To overcome this degradation in Markov-based query rewriting models, we propose a superposition-based model that continually and reactively learns locally-adaptive decision boundaries, maximizing its precision and recall over time. 
Our proposed strategies show significant improvements in self-learning tasks and overcome long-term performance degradation.
That being said, its dependence on sufficient statistical evidence for rewrite quality renders it subject to volatility with regard to tail or highly personalized rewrites, which we discuss further in the Appendix.

\bibliographystyle{acl_natbib}
\bibliography{naacl2022}

\input{sections/appendix}

\end{document}

%% file: sections/intro.tex
\section{Introduction}

Large-scale conversational AI systems such as Alexa, Google, Siri etc. serve millions of users daily all over the planet, who speak diverse languages and have a myriad of regional preferences. These models need to be constantly updated with new data to adapt to changing customer behavior and trends. Data curation processes that rely solely on human annotations cannot possibly scale to sustain the rapid update pace of these systems. 
Therefore, quite naturally, these AI agents have increased their reliance on explicit and implicit feedback from customer interactions to automate the learning process while limiting manual annotation efforts selectively only to auditing and quality control purposes.

In such feedback-based self-learning systems where new streams of data are being funneled in to continually update the system, the mere presence of the ML model itself inevitably impacts future training data. This is rather evident with query rewriting models where the reformulated query becomes intertwined with the original utterance to the extent where the successive feedback in the customer-system interaction paths become contingent on the rewrite. Here, we show that as these models continue to be updated without accounting for this unintended interference, they tend to learn false equivalencies between the original requests and rewrites, thereby impeding their own self-learning capabilities.

In this work, we build upon an absorbing Markov Chain model to make the model self-aware i.e. it can distinguish between customer requests and system rewrites, and adapt its decision boundary based on the quality of the rewrites. Note that the system can also be an ensemble of heterogeneous agents proposing different reformulations for the same query. The self-learning Markov model does not require any agent specific information and rather treats them all as a single entity. Thus, this work can be integrated into any conversational AI system to enable self-learning at a system-level without major changes to the rest of the architecture.

%% file: sections/related.tex
\section{Related Work}

Query rewriting techniques, particularly in the form of suggestive disambiguation have been extensively employed in online search systems \citep{rw:web1, rw:web2, rw:web3, rw:web4}, so as to increase recall and improve click-through rates. Naturally, conversational AI systems have also adopted similar techniques to reduce customer defects \citep{rw:mondegreen, rw:qr4, rw:qr5, rw:qr6, rw:qr1, rw:qr2, rw:qr3}. To the best of our knowledge, none of them address feedback issues that arise from model-in-the-loop environments.

Previous work has analyzed biases and noises in the feedback loop of machine learning models, particularly in recommendation systems \citep{rw:bias1, rw:bias2, rw:bias3, rw:bias4, rw:bias5, rw:bias6}. \citet{rw:loop1, rw:loop2, rw:loop3} delve into the effects of unwanted feedback loops that can lead to AI system instability. These works do not consider misplaced attribution of the feedback itself, which is exacerbated in query-rewriting systems.

In \citet{baseline}, customer interactions are modeled as an absorbing chain Markov model, and the candidate that is most likely to result in a successful absorbing state is predicted as the rewrite. This work does not address the equivalence conflation problem that occurs over time in such a setup. We update the Markov formulation to enable self-awareness and resolve the ambiguity in feedback attribution.

In \citet{rw:alibaba}, the Markov model is leveraged as a recall layer that produces candidates which are re-ranked by a self-learning neural model that relies on negative user feedback. While there is not much information on the performance of the recall layer, their neural ranking mechanism is richly augmented with common sense and various user preferences. They do not mention any degradation of the Markov model over time but it is possible that the enriched re-ranker could be compensating for this. In contrast, our work solves the issue within the self-learning Markov model itself as opposed to deferring it to a downstream model. This has the added benefit of accelerating the rate of self-learning.

%% file: sections/dataset.tex
\section{Dataset}

To extract the chains of successive customer interactions for the eventual Graph, we first pre-process about 90 days of de-identified time-series utterance data from a representative sample of customers worldwide to construct our dataset of sessions, $\mathcal{D}$.
Here, conceptually speaking, each such session represents a time-delimited snapshot of a particular customer's conversation history. To illustrate this, consider the session in Figure \ref{fig:main}(a) that encapsulates a series of consecutive utterances which follows a customer interjecting with a ``\textit{stop}'' and following up with a rephrase of their original request to play the song ``\textit{Enemy}''. Note that in practice, to maximize the consistency of a conversational goal, the time delay between consecutive turns is heuristically bounded.

Now, while the vast majority of interactions are indeed stateless, there are those which trigger dialogs so as to solicit the user to disambiguate. This inevitably creates conversational hierarchies that span multiple turns. To ground this, consider the dialog in Figure \ref{fig:template-example}(a) where the system is unable to fulfill the initiating request without first clarifying which playlist to add the song to. To address this complexity and improve the overall intelligibility of the corresponding session, such multi-turn dialogs are abridged by connecting the initiating turn with a synthetic one as shown in Figure \ref{fig:template-example}(c). This is accomplished via template-based DAGs (\textit{the construction of which is explored with greater detail in the Appendix Section \ref{appendix:template}}) wherein the resolved entities towards the end of the corresponding dialog are passed through to generate the synthetic utterance e.g. the DAG in Figure \ref{fig:template-example}(b) is fed with ``\textbf{SongName:}escape'', ``\textbf{ArtistName:}enrique iglesias'', and ``\textbf{PlaylistName:}kacey's'' so as to surface the eventual synthesized utterance, ``\textit{add escape by enrique iglesias to kacey's playlist}''.
%

%% file: sections/model.tex
\section{Self-Aware Markov Model}

Much akin to the original formulation of the Markov model by \citet{baseline}, which we henceforth regard as our baseline, our dataset of ordered linear sequence of utterances is first projected into the hypothesis space, $H$ e.g. the utterance ``\textit{play one me}'' is mapped with the aid of the system's NLU component to the hypothesis, ``\textbf{Music}|\textbf{PlayMusicIntent}|\textbf{SongName:}one me''.    
Thereafter, they are each terminated with an absorbing state. The union of these disjoint chains tantamount to our Markov Graph, $G = (V, E)$ where $V = H \cup S$ represents the set of all transient and absorbing states respectively, while $E = V \times V$, naturally corresponds to the set of edges. In a more canonical form, the Graph can be represented via the transition matrix $\mathbf{A}$:
\begin{equation}
    \label{eq:amc}
    \mathbf{A} = \begin{bmatrix}
        \mathbf{Q} & \mathbf{S} \\
        \mathbf{0} & \mathbf{I_2}
    \end{bmatrix}
\end{equation}
\noindent where $\mathbf{Q} \in \mathbb{R}^{|H| \times |H|}$ is the sub-matrix of transition probabilities between transient states such that its $(i,j)$-th element corresponds to the probability of some source transition state, $h_i$ transitioning to some target transition state, $h_j$ in a single step or mathematically speaking, $q_{i,j} = P(h_j|h_i)$. The sub-matrix $\mathbf{S} \in \mathbb{R}^{2 \times |H|}$ refers to the immediate absorption probabilities of the corresponding transient states i.e. $\mathbf{S} = \left[ \mathbf{s}^+, \mathbf{s}^- \right]$.

Now, with $\mathbf{Q}$ being a square matrix\footnote{As every atomic chain in the Graph is terminated with an absorbing state, these terminal states are guaranteed to always be reachable by any given source transient state, thus ensuring their convergence i.e. $\lim\limits_{n \rightarrow \infty} \mathbf{Q}^n = \mathbf{0}$.} whose norm, $\Vert \mathbf{Q} \Vert < 1$, the \textit{fundamental matrix} of the Markov model, $\mathbf{N}$ as formulated in Definition 11.3 by \citet{grinstead} is therefore given by $\mathbf{N} = \sum_{n=0}^{\infty} \mathbf{Q}^n =\left(\mathbf{I}_{|\mathcal{H}|} - \mathbf{Q} \right)^{-1}$
 where $\mathbf{Q}^n$ refers to the transition probability sub-matrix $\mathbf{Q}$ after exactly $n$ steps. The \textit{fundamental matrix}, $\mathbf{N}$ is leveraged in resolving the Markov model so as to surface rewrite candidates. Specifically, for a given initial transient state, $h_i$, a particular target transient state, $h_t$ would be classified as a potential candidate should it be both \textit{reachable} by $h_i$ and conditioned on $h_i$, it leads to a higher chance of success. Mathematically speaking, this optimization objective can be expressed as $\Phi_\infty (h_t) > \Phi_\infty(h_i)$ 
where $\Phi_k(h_j)$ refers to the probability of reaching a successful absorbing state, $s^+$ from $h_i$ via another state $h_j$ that is at most $k$ hops away i.e.:
\begin{equation}
    \label{eq:phi_k}
    \Phi_{k} (h_j) = P(s^+|h_j) \cdot \mathbf{N}_{i, j}
\end{equation}
\begin{figure}[t]
    \centering
    \includegraphics[width=0.48\textwidth]{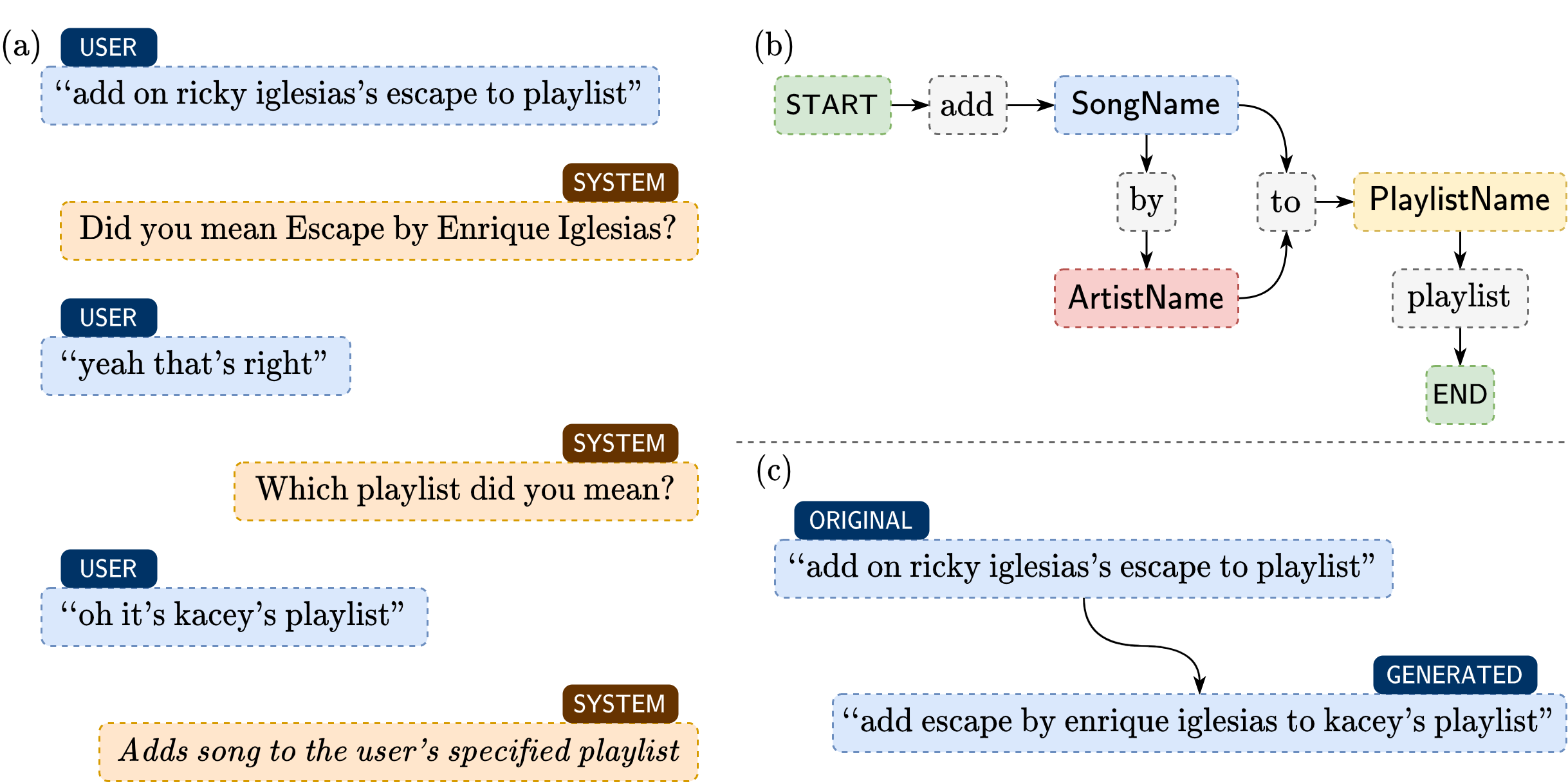}
    \caption{Dialog abridging via template-based DAG with (a) being the original dialog, (b) the extracted template graph, and (c) original with the synthesized utterance.}
    \label{fig:template-example}
\end{figure}
Here, by identifying the initial transient states that have at least one relatively more successful target transient state and thereby learning a measure of equivalency between states in the hypothesis space, $H$, the model is effectively able to partition $H$ into those that require reformulation i.e. the defective sub-space, $H^-$ and those that don't i.e. the successful sub-space, $H^+$. This nature of automatic partitioning leads the model to predict \textit{rewritability} $\hat{y}$ of a given $h_i$ as follows: 
\begin{equation}
    \label{eq:yhat}
    \hat{y}(h_i) = \mathbbm{1} \bigg\lbrace \left(\argmax_{h \in H}{\Phi_\infty(h)} \right) \neq h_i \bigg\rbrace
\end{equation}
%
\subsection{Decision Boundary Degeneracy}
Upon deployment however, the very presence of rewrites can significantly destabilize the Graph and impair the integrity of its learned partitioning. To ground this, consider, in the absence of any rewrite, a commonly misrecognized utterance, \textit{"play theme"} ($u_1$) is followed up with rephrases of \textit{"play team"}, \textit{"play the song team by lorde"}, etc. Now, when the first Markov model $G_d^{(0)}$ is trained initially at $T_0$ (Figure \ref{fig:main}b), it learns to rewrite $u_1$ to \textit{"play team by lorde"} ($r_1$). Once deployed, as the Markov model continually learns from customer feedback, $u_1$ becomes more and more successful than it actually is, since $r_1$ is not explicitly modeled. Conceptually, this \textbf{deterministic discounting} deforms the decision boundary around $u_1$, resulting in a Type II error (Figure \ref{fig:main}c). Such a misclassification will eventually shed the rewrite, forcing the graph to revert to $G_d^{(0)}$. This increases the rephrases to $u_1$ as previously observed at $T_0$ and as it gathers sufficient defect statistics, the pattern would repeat, resulting in an unstable oscillatory system that struggles to maintain a consistent decision boundary.

One way of solving the above problem, is to account for rewrites by always including them in the original interaction chain. While this might alleviate the Type II error described above, we show that this limits the system's capability to handle defective rewrites. Imagine a case where a successful utterance, say \textit{"play la da dee"} is followed up by a defective system rewrite \textit{"play lady"} (Figure \ref{fig:main}d). This may arise due to a number of reasons such as epistemic or systemic errors, multi-agent interaction, etc. as it is the nature of any statistical model. This process of \textbf{deterministic unrolling}, which presumes rewrites to have some degree of latent intent equivalency with the original utterance, would cause the original hypothesis to become more and more defective than it actually is, resulting in a Type I error. To recover the original intent, the customers would need to rephrase following the defective rewrite e.g. \textit{"play la da dee by cody simpson"} or some external guardrail mechanism would need to intervene. Yet again, the Graph will be slow to adapt the decision boundary in response to a Type I error or even worse, may completely fail to recover.

\subsection{Meta-State Augmentation}

A natural way to balance out these Type I and II errors and thereby maximizing the eventual precision and recall of the rewrites would to be to learn to unroll the rewrite should it improve the customer experience and discount it otherwise. This form of adaptive preservation and suppression of rewrites gives rise to a probabilistic decision making process where the rewrites act as a kind of meta-states that induce stochasticity within the Graph. Conceptually speaking, this is equivalent to both $G_d$ and $G_u$ being in a state of superposition as shown in Figure \ref{fig:main}(e) where in the event that a particular transient state, $h_i$ is both rewritten to $h_k$ and followed-up by $h_j$, a meta-state triplet (MST) is formed. In more robust terms, each of these MSTs within the Graph are comprised of a \textit{viability} edge, $(h_i, h_k)$, a \textit{succeeding} edge, $(h_k, h_j)$, and a \textit{discounting} edge, $(h_i, h_j)$ and are uniquely parameterized by their own set of probabilistic values, namely in this case, $\boldsymbol{\alpha}_{ik}$, $\boldsymbol{\beta}_{kj}$, and $\boldsymbol{\gamma}_{ij}$ respectively so as to allow the Graph to truly be locally adaptive in its learning. To that extent, we first construct a superposition-based transition matrix $\mathbf{A}^\prime$ by updating the probabilities as below:
\begin{equation}
    \label{eq:lambda}
    \begin{gathered}
        \mathbf{A}^{\prime} = (\boldsymbol{\lambda} \circ \mathbf{C})^\top \mathbf{D}^{-1} \\
        \boldsymbol{\lambda} = \boldsymbol{\alpha} \circ \mathbf{J}^{(\alpha)} + \boldsymbol{\beta} \circ \mathbf{J}^{(\beta)} + \boldsymbol{\gamma} \circ \mathbf{J}^{(\gamma)} + \mathbf{J}^{(\epsilon)}
    \end{gathered}
\end{equation}
\noindent where $\mathbf{C} \in \mathbb{Z}_{0+}^{|V| \times |V|}$ such that $\mathbf{C}_{xy}$ refers to the co-occurrence count of the directed edge $e_{xy} = (h_x, h_y)$ in the superposition Graph, $G^{\prime}$ and $\mathbf{D}$ is the diagonal matrix whose entries are row-wise sum of the matrix $\mathbf{C}$ i.e. $\text{diag}(\mathbf{D}) = (\boldsymbol{\lambda} \circ \mathbf{C}) \cdot \mathbf{1}$. The entries $\mathbf{J}^{(\alpha)}_{xy}, \mathbf{J}^{(\beta)}_{xy}$ and $\mathbf{J}^{(\gamma)}_{xy}$ on the other hand, are the ratios of $e_{xy}$ occurring as either a \textit{viability}, \textit{succeeding} or \textit{discounting} edge respectively. $\mathbf{J}^{(\epsilon)}_{xy}$, however, is the complementary ratio of $e_{xy}$ not being a part of any MST. As a matter of completeness, it's worth noting here that $\boldsymbol{\alpha}, \boldsymbol{\beta}, \boldsymbol{\gamma}, \mathbf{J}^{(\cdot)} \in [0, 1]^{|V| \times |V|}$ such that $\mathbf{J}^{(\alpha)}_{xy} + \mathbf{J}^{(\beta)}_{xy} + \mathbf{J}^{(\gamma)}_{xy} + \mathbf{J}^{(\epsilon)}_{xy} = 1$. Consequently, this modified transition matrix is then used in resolving the Markov Graph as before, to generate rewrite candidates.

\subsection{Meta-State Triplet Parameters}
In order to adaptively preserve or suppress the rewrites, the weights on the \textit{viability} edges, $\alpha$ should reflect the performance of rewriting. As such, for a given \textit{viability} edge $e_{xy}$ we compare the interaction quality (IQ), as scored by a neural dialog model \cite{metrics:2} of the population where $h_x$ was not rewritten, $X$ against that where $h_x$ was rewritten to $h_y$, $Y|X = W$. Now, suppose that the probability of success in each of these populations follows Beta distributions i.e. $p_X \sim \text{Beta}(a_x, b_x)$ and $p_W \sim \text{Beta}(a_w, b_w)$. Then, leveraging the beta bi-variate hypothesis testing model as formalized by \citet{bayesian-ab}, the probability that rewriting is comparatively better is given by:
\begin{equation*}
    \begin{gathered}
        P(p_W > p_X) = 1 - \int_{0}^{1} f(p_X, p_W) \cdot d_{p_X} \\
        f(p_X, p_W) = \frac{p_X^{a_x - 1}(1-p_X)^{b_x - 1}}{B(a_x, b_x)} \cdot I_{p_X}(a_w, b_w)
    \end{gathered}
\end{equation*}
\noindent where, $B$ is the beta function and $I$, the regularized incomplete beta function. Thereafter, $\boldsymbol{\alpha}_{xy}$ is computed as a variant of $P(p_X > p_Y)$ by leveraging different probability arguments depending on support sufficiency for both $p_X$ and $p_Y$ as detailed in the appendix.

Then, while $\alpha$ reflects the rewrite quality via historical statistics, the weights on the \textit{succeeding} edge, $\beta = \alpha^\rho$ are designed to maintain the semantic connectivity between the rewrite and the succeeding states. Here, we rely on Levenshtein ratio to score on both the grapheme and phoneme levels so as to compute a relevance measure, $\rho \in \mathopen[0, 1\mathclose]$. Intuitively speaking, it allows the $\alpha$-$\beta$ flow to be dampened in the event the rewrite is followed up with a semantically similar rephrase, indicating that it may not have quite achieved the customer's true intent. In a complementary fashion, the weight of the \textit{discounting} edge $\gamma = 1 - \alpha \cdot \beta$ acts as a response whose magnitude correspond to how much the corresponding rewrite in its MST needs to be suppressed. Thus, the locally adaptive Markov model is \textbf{self-aware} to be able to tailor the decision boundary so as to surgically maximize the precision and recall over the space of rewrites.

%% file: sections/experiments.tex
\section{Experiments}
We build an evaluation dataset of request-rewrite pairs annotated by a cascaded labeling pipeline comprising of an interaction quality model, NLU scores and manual verification. This fundamentally enables us to surface, for a given request, $u$, both the set of rewrites which significantly improve the customer experience, $\mathbf{r}_u^+$ and the set that significantly worsen, $\mathbf{r}_u^-$ to collectively yield our core evaluation dataset, $\mathcal{D}_e$. Then, for any given request, we further define its \textit{rewritability}, i.e. a binary label which indicates whether a particular request, $u$, should at all be rewritten, as $y_u = \mathbbm{1}(|\mathbf{r}_u^+| > 0)$.

We benchmark our self-aware Markov model variant $\mathcal{M}_{s}$ against the baseline $\mathcal{M}_{b}$ \citep{baseline} \footnote{ To the best of our knowledge, this is a novel space where widely peer-reviewed work on continual adaptive self-learning systems are few and far between. As such, this Markov-based baseline which has already shown to outperform a pointer-generator LSTM is chosen given its already established production impact.} and measure the gains introduced by our template-based generation strategy on both model variants, denoted by the subscript $+g$. Specifically, we measure their performance on the evaluation set $\mathcal{D}_e$ over three tasks, namely their ability to \textbf{partition} the requests based on their predicted \textit{rewritability}, learn the optimal rewrite for a given request i.e. \textbf{equivalence learning}, and react to changing customer preferences i.e \textbf{reactivity rate}.

\subsection{Partitioning}
The automatic partitioning task is a binary classification problem where the ground truth label $y_u$ is compared against the model prediction (Equation \ref{eq:yhat}). We observe that the self-aware models significantly improve precision and recall compared to their baseline counterparts as shown in Table \ref{table:partition}.
\begin{table}[h!]
\centering
\resizebox{\linewidth}{!}{
    \begin{tabular}{ l|c|cc } 
        \toprule
        Model & $\mathcal{M}_{b+g}$ & $\mathcal{M}_s$ & $\mathcal{M}_{s+g}$ \\
        \midrule
        Precision &  +0.0961 & $\mathbf{+0.1808}$ & +0.1688 \\
        Recall    &  +0.1724 & +0.4674 & $\mathbf{+0.5110}$ \\
        Accuracy  &  +0.0606 & +0.1922 & $\mathbf{+0.2047}$ \\
        $F_1$     &  +0.2555 & +0.5547 & $\mathbf{+0.5834}$ \\
        \bottomrule
    \end{tabular}
}
    \caption{Partitioning metrics measured as improvement over $\mathcal{M}_{b}$ }
    \label{table:partition}
\end{table}
Here, it is worth mentioning that the consistent significant gain in recall with template-based generation enabled is in part due to a strong correlating property between the need for rewriting and the need for disambiguation, which otherwise would have been lost due to the local Markov property.

\subsection{Equivalence Learning}
\begin{figure}[h]
    \centering
    \includegraphics[width=0.48\textwidth]{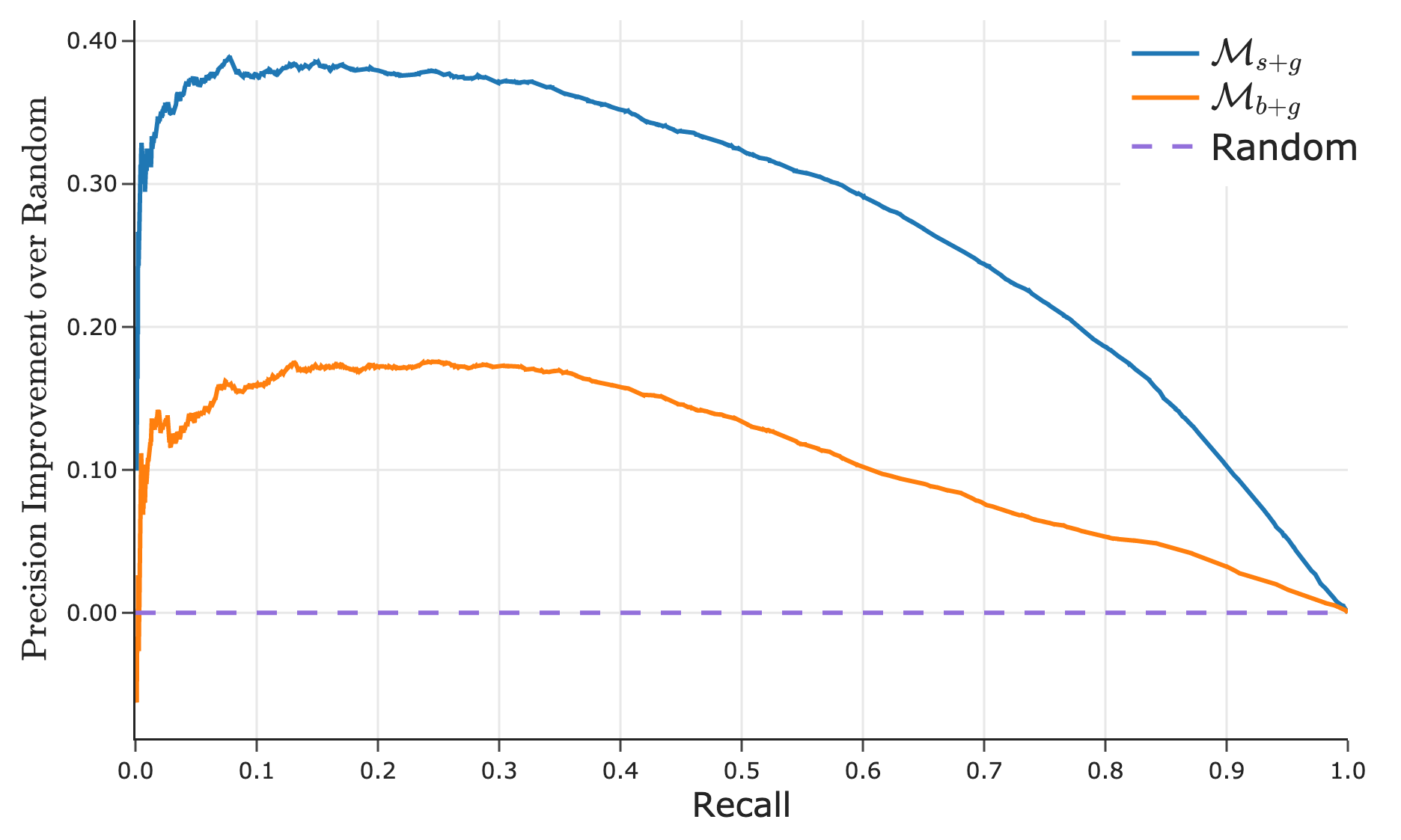}
    \caption{Precision-Recall Characteristics of Equivalence Learning.}
    \label{fig:exp-pr-curve}
\end{figure}

Once the requests are partitioned, the performance of the model in selecting rewrites i.e. its ability to optimally learn \textit{equivalencies} for those in $\mathcal{H}^-$ are evaluated. To this end, we compare the score of the models ($\Phi_\infty$ from Equation \ref{eq:phi_k}) against the ground truth annotations in $D_e$ i.e. whether a given rewrite candidate makes the customer experience significantly better (+1) or worse (-1). The precision-recall curves are then obtained as in Figure \ref{fig:exp-pr-curve}. The self-aware models exhibit much better precision vs. recall trade-offs and have significantly higher areas under the curve. To highlight, the template augmented self-aware model $\mathcal{M}_{s+g}$ improves the PR-AUC by \textbf{27.45\%} relative to $\mathcal{M}_{b+g}$.
\subsection{Reactivity Rate}
\begin{figure}[h]
    \centering
    \includegraphics[width=0.48\textwidth]{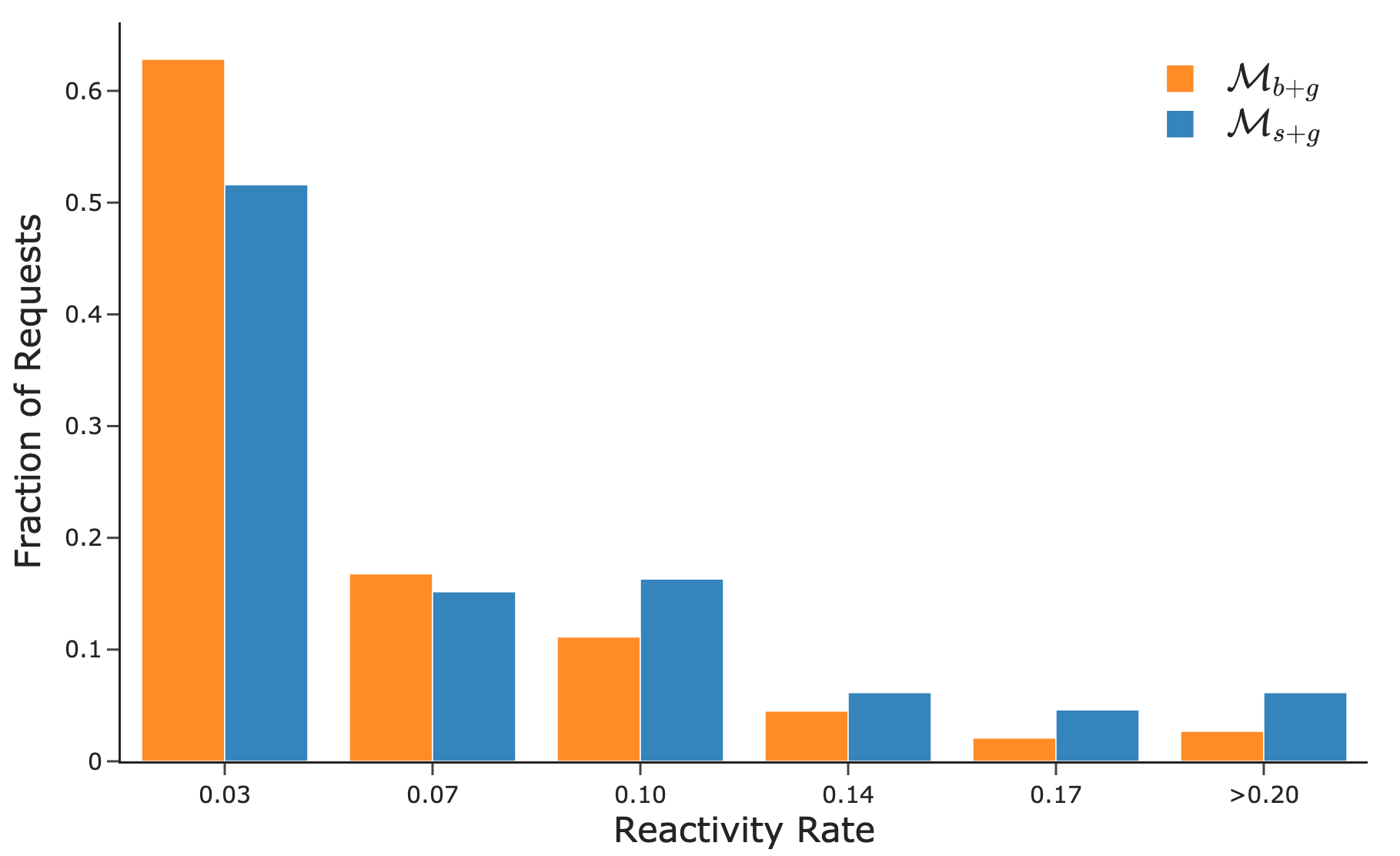}
    \caption{Reactivity Rate Distribution.}
    \label{fig:exp-reactivity}
\end{figure}

\begin{figure}[h]
    \centering
    \includegraphics[width=0.48\textwidth]{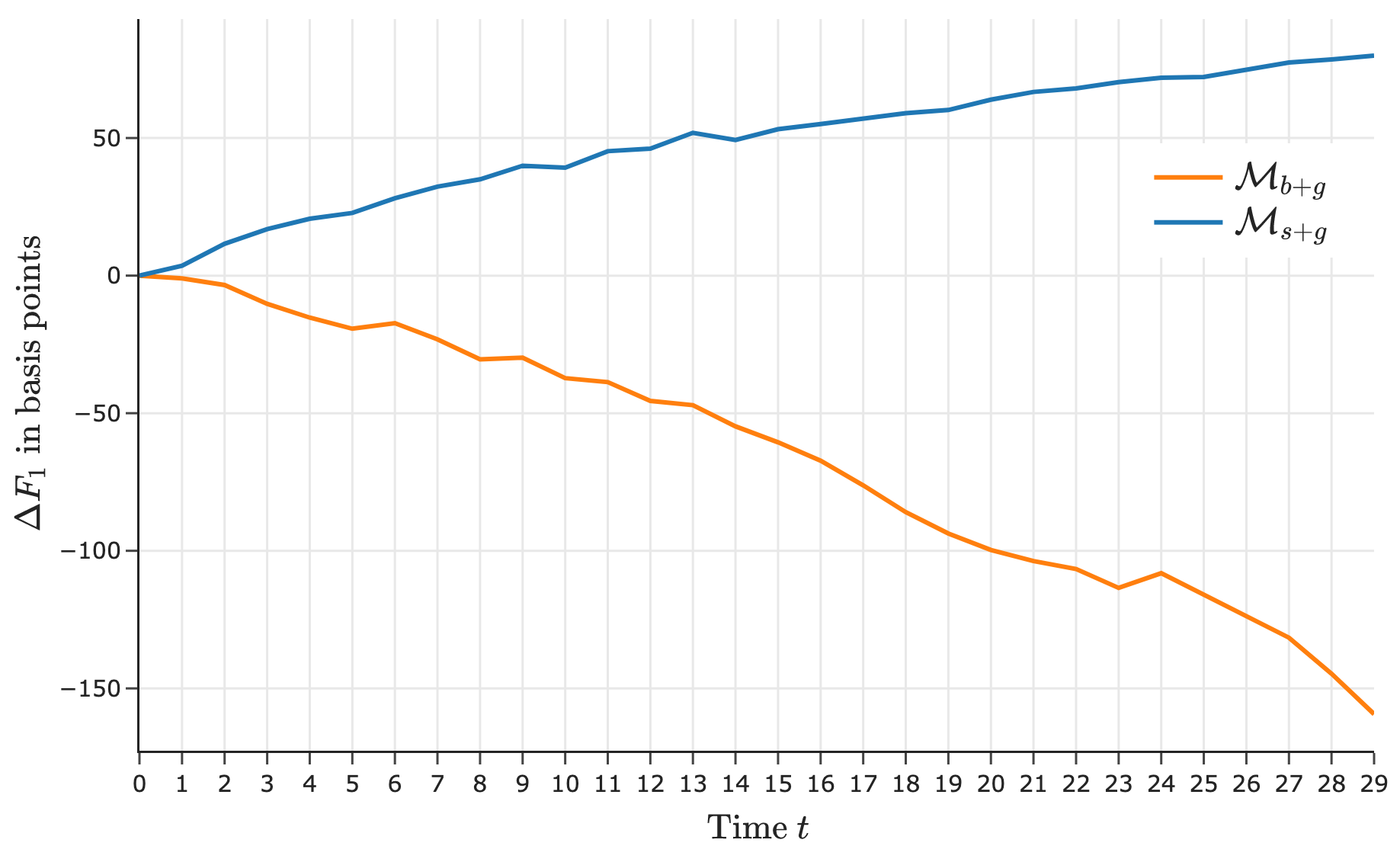}
    \caption{ 
        Relative change in $F_{1}$ score over time $t$. Note that for every timestamp, both models were retrained with new customer feedback.
    }
    \label{fig:exp-f1-trend}
\end{figure}

A key paradigm in designing large-scale AI solutions is the adaptability of the system to changing customer preferences. In the query rewriting domain, this quality can be expressed via the rate at which the top rewrite candidate changes over time i.e. the \textit{reactivity rate}. Figure \ref{fig:exp-reactivity} shows the distribution of reactivity rate for common requests across the graph over a 30 day time period. 
The self-aware model exhibits higher reactivity as seen by the right shift in the distribution with respect to the baseline. To study the impact on performance over time, we compare the relative change in $F_1$ scores of the models $\Delta F_{1}^{(t)} = \frac{F_{1}^{(t)}}{F_{1}^{(0)}} - 1$ where, $F_{1}^{(t)}$ is the $F_{1}$ score of the given model at a given timestamp $t$ on the equivalence learning task. It can be seen from Figure \ref{fig:exp-f1-trend} that the self-aware model shows relative increase in the score over time, whereas the baseline is subject to a degradation in performance. Thus the higher reactivity rate of self-awareness is correlated to increased self-learning with the models adapting to customer feedback.

\subsection{Online Performance}

With our approach for template-based generation being inherently scalable across languages and our self-aware Markov Graph naturally being language agnostic, we successfully deployed the model across 11 locales spanning 6 languages worldwide. To facilitate the models' ability to be continually adaptive, they are refreshed daily with new customer feedback. After nearly 6 weeks of in-depth A/B testing in production, we observed a strongly significant reduction (i.e. achieving a $p$-value of $\leq 0.0001$) in defects experienced by the customers compared to the baseline (see Table \ref{table:online}) with a relative defect reduction of up to \textbf{31.22\%}.

\input{sections/appendix/online}

%% file: sections/appendix/online.tex
\begin{table*}[h]
\centering
\resizebox{\textwidth}{!}{
\begin{tabular}{ l|c||cl }
    \toprule
    \textbf{Language}
    & \textbf{Defect Reduction}
    & \textbf{Example Request}
    & \multicolumn{1}{c}{\textbf{Example Rewrite}} 
    \\
    \midrule
    
    \multirow{2}{*}{English} 
    & \multirow{2}{*}{$25.78\%$}
    & \multirow{2}{*}{play tokyo take out} 
        & \textsc{old}: play tokyo \textbf{takedown}
    \\
    &
    &
        & \textsc{new}: play \textbf{towkyo takeout by michael giacchino}
    \\
     \midrule
    
    \multirow{2}{*}{French} 
    & \multirow{2}{*}{$31.22\%$}
        & mets la chanson le
        & \textsc{old}: mets \textbf{le} dimanche \`a bamako
    \\
    &
        &  dimanche \`a bamako
        & \textsc{new}: \textbf{joue la album dimanche} \`a bamako \textbf{par amadou}
    \\
    \midrule
    
    \multirow{2}{*}{Italian} 
    & \multirow{2}{*}{$23.98\%$}
    & metti campioni del mondo
        & \textsc{old}: \textbf{metti la} canzone campion\textbf{i} del mondo
    \\
    &
    &  
        & \textsc{new}: \textbf{riproduci} canzone \textbf{italia} campion\textbf{e} del mondo \textbf{di gigione}
    \\
    \midrule
    
    \multirow{2}{*}{German} 
    & \multirow{2}{*}{$22.73\%$}
    & spiel sun goes down von lenas x.
        & \textsc{old}: spiel sun goes down von lil nas \textbf{you}
    \\
    &
    &  
        & \textsc{new}: spiel sun goes down von lil nas \textbf{x}.
    \\
    \midrule
    
    \multirow{2}{*}{Spanish} 
    & \multirow{2}{*}{$28.06\%$}
        & reproducir feliz cumpleaños
        & \textsc{old}: \textbf{pon las mañanitas con} alejandro fern\'andez
    \\
    &
        &  de alejandro fern\'andez
        & \textsc{new}: reproduc\textbf{e las mañanitas} de alejandro fern\'andez
    \\
    \midrule
    
    \multirow{2}{*}{Portuguese}
    & \multirow{2}{*}{$26.21\%$}
    & \multirow{2}{*}{toca mulher chorona} 
        & \textsc{old}: toca mulher chorona de \textbf{corpo e alma}
    \\
     &
     &
        & \textsc{new}: toca\textbf{r} mulher chorona de \textbf{trio parada bruta}
    \\
    
    
    \bottomrule
    \end{tabular}
}
    \caption{Online Performance of $\mathcal{M}_{s+g}$ with Qualitative Examples.}
    \label{table:online}
\end{table*}

%% file: sections/deployment.tex
\section{Deployment}

In similar fashion to the well-established architecture of modern conversational AI systems \citep{conversational-ai}, Alexa follows suit in which the user-spoken audio is first transcribed into an utterance text by an automatic speech recognition (ASR) system and thereafter has its domain, intent and entities inferred by the natural language understanding (NLU) system. However, with the presence of our reformulation engine as shown in Figure \ref{fig:prod} below, the utterance text is intercepted so as to vend out a rewrite by means of an online database-backed lookup system before being funneled through to NLU. Thereafter, the resulting interpretation in context of the active dialog is leveraged to execute the corresponding action and respond back to the user.

\begin{figure}[h]
    \centering
    \includegraphics[width=0.47\textwidth]{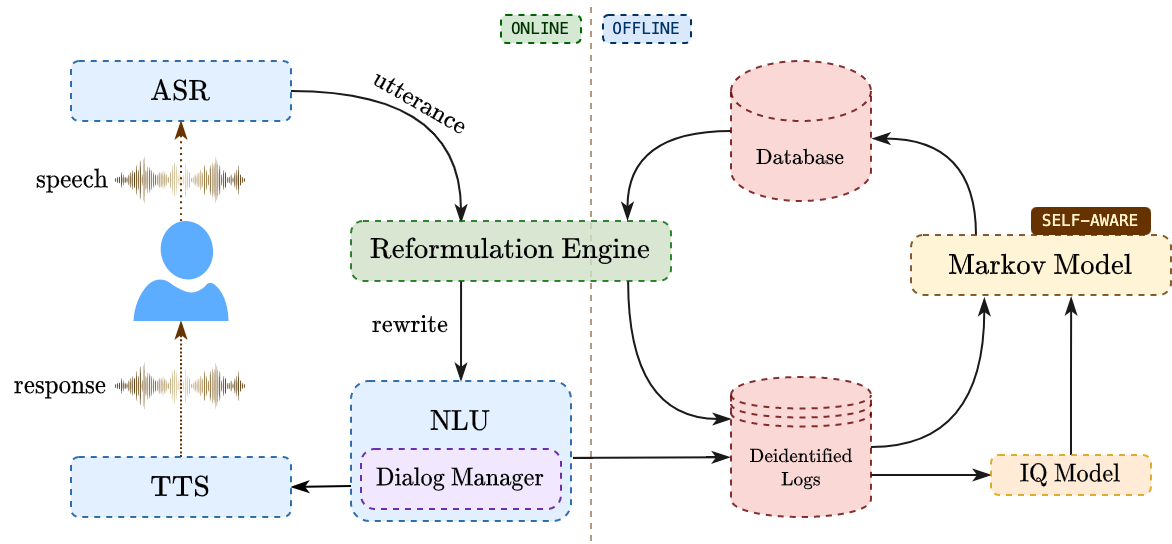}
    \caption{System Architecture}
    \label{fig:prod}
\end{figure}

Within the offline data cycle, the de-identified logs are enriched with defect predictor labels by the interaction quality (IQ) model before being collectively used to train the self-aware Markov model. The resulting rewrites surfaced by the Markov model are successively uploaded to the aforementioned online database. It is worth noting here that the offline data cycle in entirely is executed on a daily cadence so as to ensure the overall reactivity of the system. In contrast to the baseline Markov Graph, training the self-aware model incurs a rather moderate ($\sim 8.33\%$) computational overhead due to the additional $\alpha$ computation and the increased amount of edges.

%% file: sections/appendix.tex
\section{Appendix}
\label{sec:appendix}

\input{sections/appendix/template}

\input{sections/appendix/alpha}

\input{sections/appendix/limitations}

%% file: sections/appendix/template.tex
\subsection{Template-Based Generation}
\label{appendix:template}

%
%

While most interactions are single-turn, i.e. closed-form requests that are information complete, there are nonetheless dialogs that serve to disambiguate the user's intention. Such multi-turn interactions introduce conversational hierarchies, rendering each subsequent dialog turn contextually and cumulatively dependent on all its preceding turns. To ground this, consider the pair of requests -- ``\textit{set an alarm for tomorrow}" and ``\textit{set an alarm for seven a. m.}". While the latter is informationally sufficient for the system to take the requisite action, the former in contrast remains ambiguous and warrants multiple turns. Under Markov conditions where the conditional distributions are entirely uni-variate, such hierarchies are not simultaneously observed by the model and fundamentally prevent it from providing an optimal rewrite.

\begin{figure}[h]
    \centering
    \includegraphics[width=0.449\textwidth]{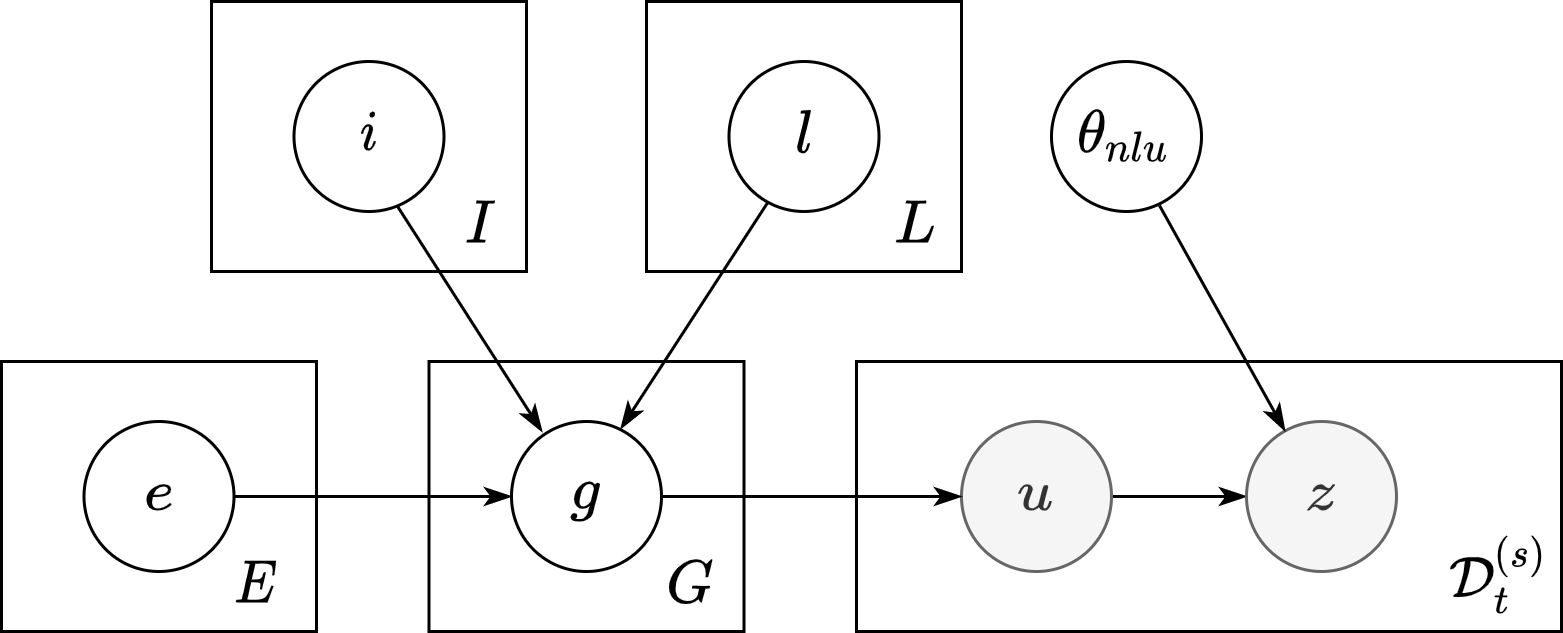}
    \caption{Plate notation summarizing the relationship between intents $I$, languages $L$, entity sets $E$, the corresponding templates $G$ and the consequent utterances and confidences in the single-turn training dataset, $\mathcal{D}_t^{(s)} = \{(u, z)^{(1)}, \dots, (u, z)^{(k)}\}$.}
    \label{fig:template-plate}
\end{figure}


\begin{figure}[h]
    \centering
    \includegraphics[width=0.47\textwidth]{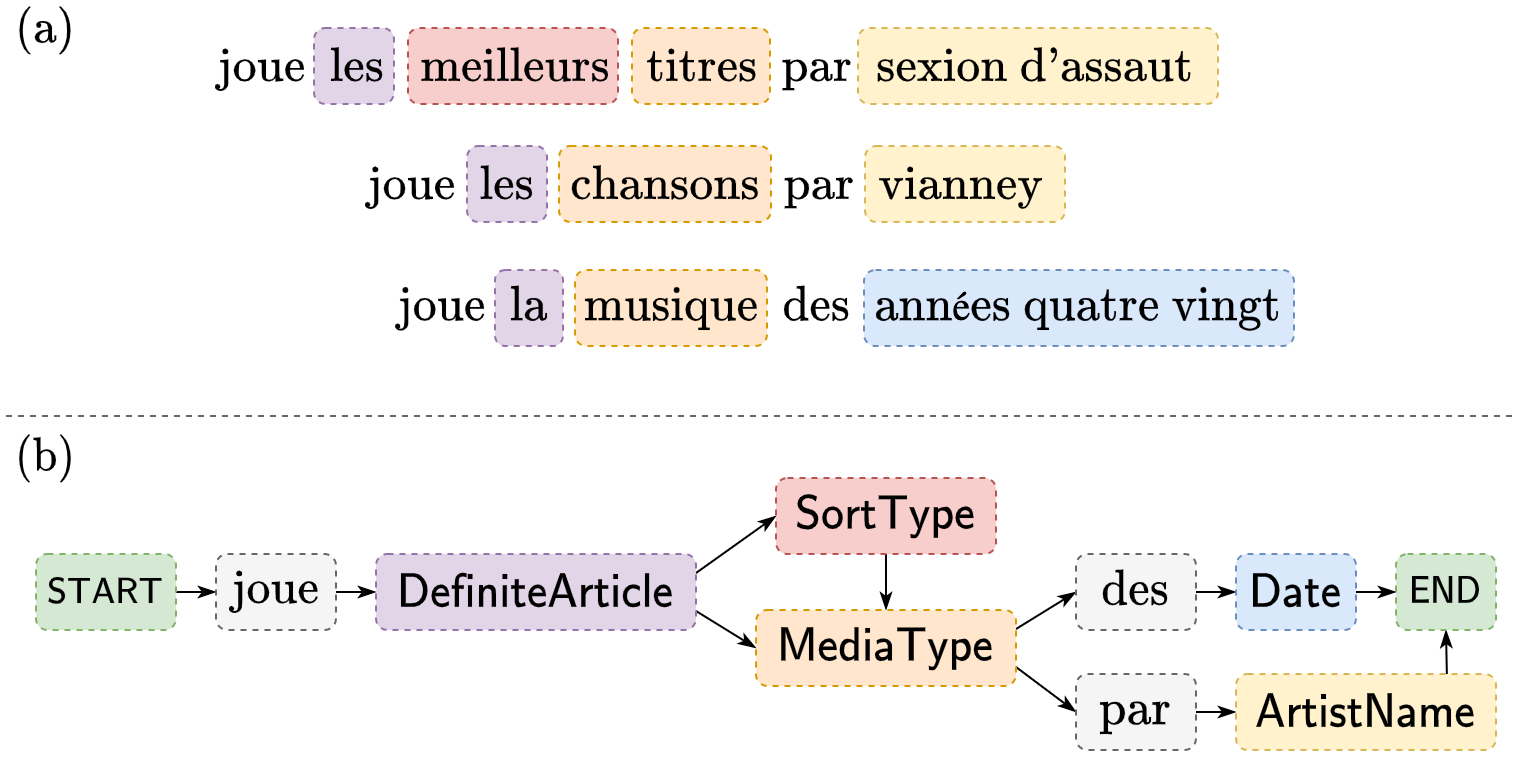}
    \caption{Template DAG extraction via NER and POS tagging with (a) showing multiple utterances with their entities and articles in colored boxes, and (b) representing the DAG for those utterances.}
    \label{fig:template}
\end{figure}

To address the limitation of the local Markov property in multi-turn dialogs, we introduce a synthetic utterance generation strategy that abridges the aforementioned hierarchy into a mere pair of turns. 
%
We define the single-turn training dataset $\mathcal{D}_t^{(s)}$ as described in the plate notation in Figure \ref{fig:template-plate}. 
We form the dataset of utterances $u$ by sampling from a distribution of templates that are conditioned on entity sets, languages, and user intents. 
These templates are obtained by leveraging NER and POS tagging results from NLU, 
as shown in Figure \ref{fig:template}a. 
Note, however, that a template $g$ leads to utterances that are not enforced to follow a proper grammatical form---potentially reflecting a low NLU confidence $z$.
Thus, for a specific entity set $e$, an intent $i$, and a language $l$, we determine the most plausible template $g^*$ by maximizing the expected value of the NLU confidence $z$:
\begin{equation}
    g^* =
        \argmax_{ 
            g \sim p_{g \mid e, i, l}
        } 
        E\mathopen[z \mid g\mathclose]
\end{equation}
where $p_{g \mid e, i, l}$ denotes the sampling probability for the template $g$ conditioned on its corresponding entity type, language, and intent. 
%
%
Once we have the set of templates for a given language and intent, we convert each template into a token chain and unify nodes across chains to form a single graph (see Figure \ref{fig:template}b). 
Although this graph is constructed from high-quality templates, it may contain cycles that prevent a proper synthetic utterance generation. 
Therefore, we factorize the graph into multiple directed acyclic graphs (DAGs). 
%
%
We identify and break cycles using depth-first search to ensure directedness while preserving the syntactic integrity of the original linguistic structures. 
This process results in multiple DAGs that account for all the original valid paths.
%

When generating synthetic utterances, we extract the entities from a multi-turn dialog and obtain the template $g^*$ that maximizes the overlap between its entity types $e$ and the DAG nodes $N_g$:
\begin{equation}
    \argmax_{g^* \in G^*_{(i, l)}} ~ | e \cap N_{g^*} |
\end{equation}
where $G^*_{(i, l)}$ is the set of optimal templates that defines the DAG and $(i,l)$ denotes a common intent and language across those templates.
Once the path has been determined, we replace the entities in template $g^*_{(i,l)}$ with their corresponding values and resolve the entity articles, if applicable.
It is possible, however, that the algorithm may not necessarily find a satisfactory path among the DAGs defined from $G^*_{(i, l)}$.
%
%
In such cases, we abridge the entire dialog to merely retain the first turn of the dialog.
Additionally, our algorithm is only executed when the multi-turn dialog has a successful conversion (i.e., the user's request was satisfied). 
In the event of an unsuccessful dialog or an abrupt end (e.g. ``\textit{no}'', ``\textit{stop}''), we terminate the dialog with an interjectory utterance.
Figure \ref{fig:template-example} describes the high-level process of compressing a multi-turn dialog into a single-turn dialog. 

%% file: sections/appendix/alpha.tex
\subsection{Meta-State Augmentation}

The weight $\alpha$ is chosen in a hierarchical fashion as follows. We select the first $\alpha$ from the successive preference relation, $\alpha_c \succ \alpha_g \succ \alpha_e$ whose confidence interval widths given by Wilson's method for both the utterance and rewrite are lesser than $\eta$. Here, the Wilson's score interval is computed with a significance of 89\% CI and $\eta$ was calibrated via cross-validation to an optimal value of $0.588$.
Each of the $\alpha_c, \alpha_g$ and $\alpha_e$ is defined by the following probability arguments,
\begin{align*}
    \alpha_c &= P(p_{W \mid c} > p_{X \mid c}) \\
    \alpha_g &= P(p_W > p_X) \\
    \alpha_e &= P(p_{W_e} > p_{X_e})
\end{align*}
where $\alpha_c$ relies on the supporting statistics for a given customer, $c$ while $\alpha_g$ extends that statistic globally across all customers in the data.
Unlike $\alpha_c$ and $\alpha_g$, however, we determine $\alpha_e$ by the distributions of entity changes between the utterance and the rewrite.  
%
Given the entity set $e$, along with their corresponding changes between the original and its rewrite (e.g., \textbf{ArtistName} added, \textbf{SongName} changed, etc.), we compute $\alpha_{e_i}$ for every entity $e_i \in e$ and retrieve the maximum absolute deviation as $\alpha_e$:
\begin{equation}
    \label{eq:alpa_e}
    \alpha_e = \max_{ e_i \in e} ~|\alpha_{e_i} - 0.5|
\end{equation}
We choose the maximum absolute deviation because it linearly provides a sense of dispersion without overly weighting values as in other formulations (e.g., standard deviation). 
More importantly, Equation \ref{eq:alpa_e} defines $\alpha_e$ based on a single most-dispersed $\alpha_{e_i}$ value, which can lead to either suppress (i.e. low dispersion) or encourage (i.e. high dispersion) the $\alpha\beta$-path. 

%% file: sections/appendix/limitations.tex
\subsection{Risks and Limitations}
\label{appendix:risks}

In order to be locally adaptive i.e. decisively unroll or discount a particular rewrite when warranted so, the learning of the Graph hinges on its ability to determine the \textit{viability} i.e. the $\alpha$ value of the said rewrite---the performance of which is squarely correlated with that of the IQ model and thereby inheriting the model's limitations in its overall precision and recall. That being said, the Graph does internally rely on its collaborative filtering ability to regularize the model's decision while external guard-rail mechanisms are also in place to further mitigate the impact of this dependency. 

Another matter of concern here would be the requisite for sufficient statistics when computing $\alpha$, which becomes a limiting factor for highly tail or personalized rewrites, where the Graph would essentially struggle to learn a consistent decision boundary given a high entropy of plausible rewrite alternatives, resulting in its equivalency learning to be entirely contingent on the more prevalent cohort within each learning cycle. In practice however, this is far from being a considerable issue as the over-arching system takes on a multi-stage hierarchical approach that permits other personalized agents to act in lieu of the Graph, while maintaining the Graph's role for its more confident set of customer cohorts. 

Conversely speaking, should there be a significantly widespread rewrite that abruptly becomes defective, the Graph would inevitably require a substantial or quite possibly, an equally voluminous source of negative feedback to counter the highly successful prior. This in turn could subject a vast number of customers to a bad experience for a considerable amount of time that ultimately drives down the engagement. As clear of a risk this is in a deployed application setting, a veritable solution here would be to adopt a sense of recency-weighting in constructing the Graph's adjacency matrix, which stands as a worthwhile future effort. In the meantime however, we rely on external gating mechanisms that refresh far more often than the Graph to aid in mitigating the overall severity of such an issue.